\documentclass[final,5p,times,authoryear]{elsarticle}

\usepackage{amssymb}
\usepackage{bm}

\journal{???}

\begin{document}

\begin{frontmatter}



\title{A Feedforward Unitary Equivariant Neural Network}


\author[label1]{Pui-Wai Ma\corref{cor1}}
\ead{Leo.Ma@ukaea.uk}
\cortext[cor1]{Corresponding Author}
\affiliation[label1]{organization={United Kingdom Atomic Energy Authority},
            addressline={Culham Science Centre}, 
            city={Abingdon },
            postcode={OX14 3DB}, 
            country={United Kingdom}}

\author[label2]{T.-H. Hubert Chan}
\ead{Hubert@cs.hku.hk}
\affiliation[label2]{organization={Department of Computer Science, The University of Hong Kong},
            country={Hong Kong}}

\begin{abstract}
We devise a new type of feedforward neural network. It is equivariant with respect to the unitary group $U(n)$. The input and output can be vectors in $\mathbb{C}^n$ with arbitrary dimension~$n$. No convolution layer is required in our implementation. We avoid errors due to truncated higher order terms in Fourier-like transformation. The implementation of each layer can be done efficiently using simple calculations. As a proof of concept, we have given empirical results on the prediction of the dynamics of atomic motion to demonstrate the practicality of our approach.
\end{abstract}



\begin{keyword}
equivariant neural network \sep feedforward neural network \sep unitary equivariant \sep rotational equivariant


\end{keyword}

\end{frontmatter}


\section{Introduction}
Neural Networks (NN) have gained popularity in many different areas because of its universal approximator property \citep{SONODA2017233}. In recent years, equivariant NN (ENN) in different architectures have been applied in various areas, such as 3D object recognition \citep{Thomas_arxiv_2018,Esteves2020}, molecule classification \citep{Weiler_NeurIPS_2018}, interatomic potential development \citep{Kondor_arxiv_2018, batzner2022}, and medical images diagnosis \citep{Muller_arxiv_2021,Winkels_2018,Worrall_2018}. 

When NNs are employed to model some physical phenomena, they should obey certain physical symmetry rules.  For example, if an NN is intended to return some potential function between particles, the output should be \emph{invariant} with respect to rotation of input particles' coordinates.  On the other hand, for an NN predicting particle movements, the output should be \emph{equivariant} with respect to rotation, i.e., if a rotation operator is applied to the input particles' coordinates, the effect is the same as applying the same rotation operator to the output.

In some works \citep{Brandstetter2022}, equivariance is achieved through data augmentation, i.e., additional training data are created by transforming existing training data (e.g., create additional copies by rotation). However, if equivariance is implemented in an NN, one can avoid the need of data augmentation, which reduces the demand on storage and improves sampling efficiency. This is especially important if one is working on data in continuous space. For example, if input data are points in Euclidean space and the output data are translational and/or rotational equivariance or invariance, it is not practical to create too many copies of data.   

Previous works have achieved equivariance via higher order representations for intermediate network layers. For example, the implementation of spherical symmetry, such as $S^2$ or $SO(3)$, can be achieved through a layer with kernel performing a 3D convolution with spherical harmonics or Wigner D-matrices \citep{Thomas_arxiv_2018,Gerken}. This is analogous to Fourier transforms in linear space. However, these kinds of implementation are computationally expensive \citep{Cobb_ICLR_2021}.

In physical systems, although they can in principle be described by physical rules, analytical methods are not always feasible when the analytical form (such as the Hamiltonian) is unknown.  On the other hand, an NN consists of many computationally simple components that can operate in parallel, and hence, they are suitable for large scale complicated simulations, as long as there is enough training data. 

\textbf{Our contributions.} We have designed a new framework for feedforward neural networks. Specifically, it has the following properties. 
  
\begin{enumerate}
    \item 
    The inputs and outputs are vectors in $\mathbb{C}^n$. They are equivariant with respect to the unitary group $U(n)$. 

    \item In each layer, in addition to a linear combination of vectors from the previous layer, we have an extra term that is a linear combination of the normalized vectors as well.  This extra term acts like the bias term in an affine transformation.
    
    \item Each layer has an activation function that acts on vectors in $\mathbb{C}^n$ and is also equivariant with respect to unitary operators.
    
    \item Equivariance is achieved in a feedforward neural network without any convolution layer. 
    
\end{enumerate}

Moreover, in Section~\ref{sec:poc}, we have performed numerical experiments on the simulation of a physical system using our ENN framework in the scenario when the rules governing the system might be unknown.

%
%

\section{Related Works}
We compare our framework with previous approaches on equivariant neural networks.


\cite{Kondor_Trivedi} proved analytically that convolutional structure is a necessary and sufficient condition for equivariance to the action of a compact group. Therefore, many works designed the architecture of their NN based on this theorem, where convolution layer is introduced. \cite{Cohen_PMLR_2016} introduced group equivariant convolution network. They used features map functions on discrete group, and so it only works with respect to finite symmetry groups. 

\cite{Cohen_ICLR_2018} considered convolution NN for spherical images through Fourier analysis using Wigner D-matrices. \cite{Kondor_NeurIPS_2018} improved the implementation using Clebsch-Gordan decomposition, where the NN is operated in Fourier space only. It avoids the need of switching back and forth between Fourier and real spaces. 

\cite{Thomas_arxiv_2018} shows if the input and output of each layer is a finite set of points in $\mathbb{R}^3$ and a vector in a representation of $SO(3)$, one can decompose this into irreducible representation through convolution with spherical harmonics and Wigner D-matrices. \cite{Esteves2020} implements exact convolutions on the sphere using spherical harmonics. It maps spherical features of a layer to the spherical features of another layers. 

Convolution using spherical harmonics is analogous to Fourier transform in signal processing. In practice, it only preserves the most significant coefficients. Error is inherently introduced due to truncated higher order terms. It is also computation demanding due to the need of performing integration or summation. 

Our newly designed feedforward neural network guarantees equivariance without any convolution layer. We should note our ENN has a structure in vector form which is different from the conventional NN structure in scalar form that was considered by \cite{Kondor_Trivedi}. Besides, our implementation is much simpler than previous works. 

\cite{Satorras} devised an equivariant graph NN (GNN) with respect to $E(n)$ operators (that include rotation, reflection and translation). Similar to our approach, it does not contain convolution layer. The input spatial coordinates are vectors. Due to the construction of a GNN, their spatial coordinates are not filtered by activation functions. Their spatial coordinates are updated through averaging with respect to neighbors. The number of nodes in each layer is restricted to be the same, where our approach is general enough to allow different numbers of nodes in different layers. In addition to the spatial coordinates on which the operators act, their GNN contains feature vectors which do not fall under the equivariant aspect. We will also discuss how to add these extra features in our approach.  

\section{Our Framework for ENNs}
\subsection{Equivariance with respect to unitary operators}

In general, given a function $\phi: \mathcal{X} \rightarrow \mathcal{Y}$
(where the domain $\mathcal{X}$ and the co-domain $\mathcal{Y}$
might be different) and a group $G$,
we assume that each element $g \in G$
induces \emph{group actions} $T_g : \mathcal{X} \rightarrow \mathcal{X}$
and $\widehat{T}_g: \mathcal{Y} \rightarrow \mathcal{Y}$ on
$\mathcal{X}$ and $\mathcal{Y}$, respectively.  Then, the function $\phi$
is \emph{equivariant} under $G$ if for all $g \in G$ and $x \in \mathcal{X}$, the following holds:
\begin{equation} \label{eq:equi}
    \phi(T_g(x)) = \widehat{T}_g(\phi(x)).
\end{equation}
Formally, a group action needs
to satisfy $T_{g_1 g_2} = T_{g_1} \circ T_{g_2}$ for all
$g_1, g_2 \in G$.

\emph{Invariant} is the special case when for all $g \in G$,
the group action $\widehat{T}_g$ is the identity function on $\mathcal{Y}$.

In this paper, we consider domains of the form $\mathbb{C}^{n \times M}$,
which we interpret as $M$ points in $\mathbb{C}^n$.  We consider the unitary group
$U(n)$, where each element corresponds to a unitary operator $\mathcal{U}$ on $\mathbb{C}^n$. The unitary group contains the orthogonal group $O(n)$ (that corresponds to rotations and reflections) and $SO(n)$ (that corresponds to rotations only).

Given a unitary operator~$\mathcal{U}: \mathbb{C}^n \rightarrow \mathbb{C}^n$, the group action
on $M$ points are defined by 
\begin{equation}
    (\mathbf{x}^{(1)}, \mathbf{x}^{(2)}, \ldots, \mathbf{x}^{(M)}) \mapsto (\mathcal{U} \mathbf{x}^{(1)}, \mathcal{U} \mathbf{x}^{(2)}, \ldots, \mathcal{U} \mathbf{x}^{(M)}). \nonumber
\end{equation}

\subsection{Structure of our ENN}
We construct a feedforward neural network with $L-1$ hidden layers. The input layer is labelled as the $0^{th}$ layer, and the output layer
is the $L^{th}$ layer. For the $(k+1)^{th}$ layer, its input is from the $k^{th}$ layer:
\begin{equation}
    \mathbf{x}_k \in \mathbb{C}^{n\times M_k},
\end{equation}
where $M_k$ is the number of vector elements of 
\begin{equation}
    \mathbf{x}_k=\{\mathbf{x}_k^{(1)},\mathbf{x}_k^{(2)},...,\mathbf{x}_k^{(M_k)}\}.    
\end{equation}
Each vector element $\mathbf{x}_k^{(\alpha)}\in \mathbb{C}^n$ is an $n$-dimensional vector. Similarly, we have the output 
\begin{equation}
    \mathbf{x}_{k+1}\in \mathbb{C}^{n\times M_{k+1}}.
\end{equation}
We define a variable
\begin{equation}
    \mathbf{y}_k = \mathbf{x}_k \mathbf{W}_k + \mathbf{e}_k \mathbf{b}_k.
\end{equation}
This definition is different from conventional feedforward NN. First, the $\mathbf{x}_k$ is a matrix and is put on the left-hand side of the weight parameter. Second, a new matrix variable $\mathbf{e}_k$ is introduced. These two changes are crucial steps to avoid the need to perform convolution. It allows the unitary operator getting out naturally from the left-hand side of $\mathbf{x}_k$ and $\mathbf{e}_k$. 

The weight and bias parameters matrices
\begin{eqnarray}
    \mathbf{W}_k\in \mathbb{C}^{M_k\times M_{k+1}}\\
    \mathbf{b}_k\in \mathbb{C}^{M_k\times M_{k+1}}
\end{eqnarray}
and
\begin{equation}
    \mathbf{e}_k = \left\{\frac{\mathbf{x}_k^{(1)}}{||\mathbf{x}_k^{(1)}||} ,\frac{\mathbf{x}_k^{(2)}}{||\mathbf{x}_k^{(2)}||},...,\frac{\mathbf{x}_k^{(M_k)}}{||\mathbf{x}_k^{(M_k)}||}\right\},
\end{equation}
where $||.||$ is the norm of an $n$-dimensional vector. Observe that for any unitary operator $\mathcal{U}$, it holds that
\begin{equation}
   ||\mathbf{x}_k^{(\alpha)}||=||\mathcal{U}\mathbf{x}_k^{(\alpha)}||.
\end{equation}

For all $\alpha\in \{1, 2,...,M_k\}$, we can obtain
\begin{equation}
    \mathbf{y}_k (\mathcal{U}\mathbf{x}_k)  = \mathcal{U}\mathbf{x}_k \mathbf{W}_k + \mathcal{U}\mathbf{e}_k \mathbf{b}_k = \mathcal{U}\mathbf{y}_k(\mathbf{x}_k),
\end{equation}
where $\mathcal{U}$ is applied element-wise on each $\mathbf{x}_k^{(\alpha)}$.

Then, we define an activation function
\begin{equation}
    \bm{\sigma}_{k+1}(\mathbf{y}_k)=\mathbf{x}_{k+1}.
\end{equation}
for the $(k+1)^{th}$ layer. The activation function acts on each vector  $\mathbf{y}_k^{(\alpha)}\in\mathbb{C}^n$ in element-wise manner.  We note  $\mathbf{y}_k$ has the same dimension of $\mathbf{x}_{k+1}$.

We shall find an activation function that satisfies the following:
\begin{equation}
    \bm{\sigma}_{k+1}(\mathcal{U}\mathbf{y}_k)=\mathcal{U}\bm{\sigma}_{k+1}(\mathbf{y}_k).
\end{equation}
This completes the construction of our feedforward equivariant neural network for unitary transformations.

Observe that each layer is equivariant with respect to unitary operators
in the sense of equation~(\ref{eq:equi}). The reason is that if we transform
the input $\mathbf{x}_k \rightarrow \mathcal{U} \mathbf{x}_k$, then its output will undergo the transformation $\mathbf{x}_{k+1} \rightarrow \mathcal{U} \mathbf{x}_{k+1}$; in this case, the unitary operator can act element-wise on both the input and the output spaces. Therefore, when we apply the group action, which is now the unitary operator $\mathcal{U}$,  on the $0^{th}$ layer input $\mathbf{x}_0$, the same operator will propagate to the final layer output $\mathbf{x}_{L}$. It means if we put $\mathbf{x}_0\rightarrow \mathcal{U}\mathbf{x}_0$, the output will become $\mathbf{x}_{L}\rightarrow \mathcal{U}\mathbf{x}_{L}$.


A possible choice of the activation function for each element can be a softsign function with a small residue, that is 
\begin{equation}
    \bm{\sigma}(\mathbf{u}) = \frac{\mathbf{u}}{1+||\mathbf{u}||} + \mathbf{u}\times a,
\end{equation}
where $a$ is a (small) scalar constant, and $\mathbf{u}\in \mathbb{C}^n$. The small residue is to avoid vanishing gradient of Loss function when $\mathbf{u}$ is large. We used this activation function in our numerical experiment.

Alternatively, one may choose the identity function, that is
\begin{equation}
    \bm{\sigma}(\mathbf{u})=\mathbf{u},
\end{equation}
which in scalar form is a popular choice of activation function for the output layer. Similarly, ReLu function and Leaky ReLu function in vector forms are also equivariant with respect to unitary operators. 

\subsection{Including local scalar features}

We can introduce extra scalar features into our ENN, in addition to vector elements. The idea is that we will increase the number of coordinates from $n$ to $n+m$, and we only consider
unitary operations that do
not change the extra $m$ coordinates.

Formally, for each input $\mathbf{x}_0^{(\alpha)}$,
we assume that 
it has $m$ corresponding scalar features which can be written as a vector $\mathbf{h}_0^{(\alpha)} = \{h_{0,1}^{(\alpha)}, h_{0,2}^{(\alpha)}, ...,h_{0,m}^{(\alpha)}\}$, we can rewrite the input vector element into 
\begin{equation}
    {\mathbf{x}'}_0^{(\alpha)}=\{\mathbf{x}_0^{(\alpha)},\mathbf{h}_0^{(\alpha)}\}  \in \mathbb{C}^{(n+m)} ,
\end{equation}
and the unit vector
\begin{equation}
    {\mathbf{e}'}_k^{(\alpha)}=\{\mathbf{e}_k^{(\alpha)},\mathbf{1}\},
\end{equation}
where $\mathbf{1}$ is a vector with $m$ elements and all equal $1$. (Observe that in the actual implementation,
we can reduce
$\mathbf{1}$ and the associated weights in the model to a single scalar bias term.)

The operator can be rewritten in matrix form such that
\begin{equation}
\mathcal{U}' = \left(
    \begin{array}{cc}
\mathcal{U} & \mathbf{0}\\
\mathbf{0} & \mathbf{I}\\
\end{array}  \right),
\end{equation}
where $\mathbf{I}$ is an identity $m\times m$ matrix. Plugging them back to equations in previous subsection, they all hold, provided that the definition of norm can fulfill, i.e.
\begin{equation}
   ||{\mathbf{x}'}_k^{(\alpha)}||=||\mathcal{U}'{\mathbf{x}'}_k^{(\alpha)}||.
\end{equation}
It essentially means we only apply the group action on part of the input vectors, and keep the features part of the vectors fixed. Features can be anything that are quantifiable, such as color, brightness, contrast, electronic charge, mass, humidity, level of pollutant, and etc.

Although at the output layer, we will get outputs ${\mathbf{x}'}_L  \in \mathbb{C}^{(n+m)\times M_L}$, the Loss function can be defined only using part of it. We also need to be careful that ${\mathbf{x}'}_L$ does not need to have the same unit or meaning as ${\mathbf{x}'}_0$. For example, if we considers a system of molecules, we may use positions as the vector elements, and charges and masses as features. It means we have vector inputs in $\mathbb{R}^{(3+2)\times M_0}$. Even we have vector outputs $\mathbb{R}^{(3+2)\times M_L}$, the prediction can be forces, atomic energy, and a dummy value that does not enter the Loss function. On the other hand, one can also add dummy input features to make each element in ${\mathbf{x}'}_0$ and ${\mathbf{x}'}_L$ longer.

\subsection{Backpropagation}
We can derive an algorithm similar to the commonly known backpropagation. The essence of backpropagation is to reuse the information of the gradient of the Loss function with respect to the elements in weight and bias parameters. First, we define our Loss function:
\begin{equation}
    L=C\left(\mathbf{T}, \bm{\sigma}_L(\mathbf{y}_{L-1})\right),
\end{equation}
where $\mathbf{T}\in \mathbb{C}^{n\times M_L}$ is the target data, and $C$ is a non-negative real value function being differentiable with respect to $\bm{\sigma}_L$. For convenient, we write a combined representation of the weight and bias parameters, such that $\mathbf{z}_k=\{\mathbf{W}_k,\mathbf{b}_k\}$. For each element in $\mathbf{z}_{L-1}$, the derivative
\begin{equation}
    \frac{\partial L}{\partial z_{L-1,pq}}=\bm{\delta}_{L-1}\frac{\partial \mathbf{y}_{L-1}}{\partial z_{L-1,pq}},
\end{equation}
where
\begin{equation}
    \bm{\delta}_{L-1}=\frac{\partial C}{\partial\bm{\sigma}_L}\frac{\partial\bm{\sigma}_L}{\partial \mathbf{y}_{L-1}}.
\end{equation}
For other layers, in general, we can write
\begin{equation}
    \frac{\partial L}{\partial z_{L-k,pq}}=\bm{\delta}_{L-k}\frac{\partial \mathbf{y}_{L-k}}{\partial z_{L-k,pq}},
\end{equation}
where 
\begin{equation}
    \bm{\delta}_{L-k-1}=\bm{\delta}_{L-k}\frac{\partial \mathbf{y}_{L-k}}{\partial \mathbf{x}_{L-k}}\frac{\partial\bm{\sigma}_{L-k}}{\partial \mathbf{y}_{L-k-1}}.
\end{equation}
This allows us to reuse the information of $\bm{\delta}_{L-k}$ in $\bm{\delta}_{L-k-1}$. However, it is different from the conventional backpropogation, where the whole derivative of the deeper layer is reused. We only use a part of the derivative in our backpropagation procedure.

\subsection{Permutation symmetry}

The above construction of ENN has no permutation symmetry yet.  We say that a function $\phi$ having $n$ inputs achieves permutation symmetry, if for any permutation $\pi$ on $\{1, 2, \ldots, n\}$ and any $(x_1, \ldots, x_n)$,
\begin{equation}
    \phi(x_1, \ldots, x_n) = \phi(x_{\pi(1)}, \ldots, x_{\pi(n)}).    
\end{equation}
The order and the number of inputs of $\phi$ are fixed. 
In some applications, 
we may partition the inputs and consider permutations
within each part.
For instance, we can partition the inputs into
$\mathcal{S}_a=\{1, 2, \ldots, m\}$ and $\mathcal{S}_b=\{m+1, \ldots, n\}$ such that for any permutations
$\pi_a$ and $\pi_b$ on the corresponding parts,
permutation symmetry means:
\begin{equation} \label{eq:sym}
    \phi(x_1, \ldots, x_n) = \phi(x_{\pi_a(1)}, \ldots, x_{\pi_a(m)}, x_{\pi_b(m+1)}, \ldots, x_{\pi_b(n)}). 
\end{equation}

Permutation symmetry should hold in some physical systems. For example, if we have a molecular composed of H, C and O atoms, then each input corresponds to an atom and the inputs
can be partitioned according to the type of atom.  Then, atoms in each part can be permuted without affecting the output.

One way to achieve permutation symmetry is
to introduce a pre-processing step in the first layer. 
Suppose there are $N_0$ input vectors denoted by
$(\mathbf{u}^{(\xi)}\in \mathbb{C}^{n}: \xi \in \{1,2,...,N_0\})$.
This step produces $M_0$ vectors via
a collection of functions
$\mathbf{D}^{(\alpha)}: \mathbb{C}^{n\times N_0}\rightarrow \mathbb{C}^{n}$
for each $\alpha \in \{1,2,...,M_0\}$.
Since our final neural network
achieves equivariance, we require that
for any unitary operator $\mathcal{U}$, the following holds:

\begin{equation}
    \mathcal{U}\mathbf{D}^{(\alpha)}(\{\mathbf{u}^{(\xi)}\})=\mathbf{D}^{(\alpha)}(\{\mathcal{U}\mathbf{u}^{(\xi)}\}).\label{UD}
\end{equation}


In addition to being equivariant for unitary operators, 
we describe how each $\mathbf{D}^{(\alpha)}$ also achieves
permutation symmetry.

\noindent \textbf{Achieving Permutation Symmetry by Summation.}  Even though the functions implemented
in a layer can look complicated, the principle behind them to achieve permutation symmetry is very simple. An example for $\phi$ in equation \ref{eq:sym} can be:
\begin{equation}
    \phi(x_1, \ldots, x_n) =  \sum_{i=1}^n x_i w,
\end{equation}
where $w$ is a trainable weight.  The key observation is that $w$ does not depend on the index~$i$, which is subject to permutation.
Hence, when the indices~$i$ are permuted,
the value of the function does not change.

We can also express the idea of partitioning
the $n$ inputs and consider permutation symmetry within each part.  For example, each part is indexed by~$\delta$ (also known as a feature), and an index~$i$ having feature~$\delta$ can be represented by $v_{i\delta} = 1$ and 0, otherwise.  Then, we can consider the following function:
\begin{equation}
    \phi(x_1, \ldots, x_n) =  \sum_{\delta} \sum_{i=1}^n x_i v_{i \delta} w_\delta,   
\end{equation}
where the trainable weights $w_\delta$ again
does not depend on the index~$i$.  Observe that
if indices~$i$ having the same feature~$\delta$
are permuted, the value of the function does not change.

\noindent \textbf{Graph Neural Network Example.}
Applying the above principles for permutation symmetry, we may consider adding a layer of GNN to our ENN. First, we can write a set of scalar functions for node $i$, \begin{equation}
    D_i^{(\alpha)}={v'}_{i\alpha}=\sigma\left( \sum_{j,\beta,\delta} e_{ij}^\beta v_{j\delta} w_{\beta\delta\alpha}\right)\label{D_i_scalar}
\end{equation}
where $\beta$ are features of edge $\mathbf{E}$, $\delta$ are features of node $\mathbf{V}$, $\alpha$ are features of node $\mathbf{V}'$, $\sigma$ is the activation function, and $w_{\beta\delta\alpha}$ is a trainable rank-3 tensor weight between features $\beta$, $\delta$ and $\alpha$. We can see such scalar functions hold permutation symmetry, and are rotational invariant. However, they are not in vector equivariant form. We may resolve the issue by devising a set of equivariant vector functions
\begin{equation}
    \mathbf{D}_i^{(\alpha)}={\mathbf{v}'}_{i\alpha}=\bm{\sigma}\left( \sum_{j,\beta,\delta} \frac{\mathbf{u}_{ij}}{u_{ij}}e_{ij}^\beta v_{j\delta} w_{\beta\delta\alpha}\right),
\end{equation}
provided that $e_{ij}^\beta$ is invariant with respect to the application of group action on $\mathbf{u}_i$ and $\mathbf{u}_j$, where $\mathbf{u}_{ij} = \mathbf{u}_j - \mathbf{u}_i$. The vector activation function $\bm{\sigma}$ is also required to be a vector equivariant function.

If we now consider a system of atoms, for atom $i$, it has $N_0$ neighbors within a cutoff distance $r_c$. $\{\mathbf{r}_j\in \mathbb{R}^3 | r_{ij} < r_c\}$ is the set of positions of neighboring atoms of atom $i$. We may consider each atom as a node. As a special case, one can put
\begin{equation}
    e_{ij}^\beta= \left\{
    \begin{array}{lc}
         \exp (-\eta^{(\beta)} (r_{ij}-r_s^{(\beta)})^2) f_c(r_{ij}), & \textrm{if } i\neq j\\
         0, & \textrm{if } i=j
    \end{array}\right.
\end{equation}
where the interatomic distance between atom $i$ and $j$ is $r_{ij}=|\mathbf{r}_{ij}|=|\mathbf{r}_j - \mathbf{r}_i|$. The $f_c$ is a scalar smooth-out function such that at the cut-off distance $r_c$, $f_c(r_c)=0$, and is continuous and differentiable up to at least second derivatives. This is similar to the implementation in SchNet \citep{SchNet}.

We further assume there is only one feature corresponding to $\delta$, where $v_{j1}=1$, the weight is a Kronecker delta function $w_{\beta 1\alpha}=\delta_{\beta\alpha}$, and the activation function is an identity function. Equation \ref{D_i_scalar} becomes:
\begin{equation}
    D_i^{(\alpha)} (\{\mathbf{r}_j\}) = \sum_{j,j\neq i} \exp (-\eta^{(\alpha)} (r_{ij}-r_s^{(\alpha)})^2) f_c(r_{ij}),\label{BP_descriptor}
\end{equation}
where $\alpha\in\{1,2,...,M_0\}$. The set of hyper-parameters $\{\eta^{(\alpha)}, r_s^{(\alpha)}\}$  are predetermined values. Interestingly, this is in the same functional form as suggested by \cite{BP_prl_2007} who mapped the local atomic environment to a set of atom-centered symmetry functions (or called spatial descriptors) and used them to develop machine-learned (ML) interatomic potential. 

Following similar logic, one may devise a set of vector function by augmenting above scalar spatial descriptor, such that
\begin{equation}
    \mathbf{D}_i^{(\alpha)} (\{\mathbf{r}_j\}) = \sum_{j,j\neq i} \frac{\mathbf{r}_{ij}}{r_{ij}}\exp (-\eta^{(\alpha)} (r_{ij}-r_s^{(\alpha)})^2)f_c(r_{ij}).
\end{equation}
It is straightforward to check
\begin{equation}
    \mathcal{U}\mathbf{D}_i^{(\alpha)} (\{\mathbf{r}_j\})=\mathbf{D}_i^{(\alpha)} (\{\mathcal{U}\mathbf{r}_j\}),
\end{equation}
which resembles equation \ref{UD}. The subscribe $j$ here is corresponding to the superscribe $(\xi)$ of $\mathbf{u}$ in equation \ref{UD}.

We are not going to show any example on this. It is only to show theoretically that one can introduce permutation symmetry by adding an extra layer of properly designed vector form GNN to our ENN.

\subsection{Limitations of our ENN}
Our implementation has the limitation that the group action on the input data is the same group action on the output data, and the group action is restricted to unitary transformation. Therefore, if one applies the unitary operator on the input data, but the target data does not experience the same transformation, our method does not apply. 

For example, in physical systems, quantities can have odd or even parity symmetry, e.g.,
consider the parity transformation $\mathcal{P}:(x,y,z)\mapsto(-x,-y,-z)$. For quantities with odd parity symmetry, they will have sign change according to the parity transformation. Our ENN can be applied to predict these quantities. However, for vector quantities with even parity symmetry, our ENN does not apply. For example, in classical mechanics, the angular momentum
\begin{equation}
    \mathbf{L}=\mathbf{r}\times\mathbf{p}.
\end{equation}
If we apply the parity transformation, we get $\mathbf{r}\rightarrow\mathbf{-r}$ and $\mathbf{p}\rightarrow\mathbf{-p}$, but we still get
\begin{equation}
    \mathbf{L}=\mathbf{-r}\times\mathbf{-p}.
\end{equation}
If we use $\mathbf{r}$ and $\mathbf{p}$ as the input data, we cannot use our ENN to predict $\mathbf{L}$. A possible solution is to manually apply sign change to the output data according to the input data.

For scalar quantities with even parity symmetry, it can be remedied by converting $\mathbf{x}_k$ at arbitrary $k$ layers, where $\mathbf{x}_k\in\mathbb{C}^{n\times M_k}$, to invariant scalar quantities.
For examplem we can set $\mathbf{w}_p$ as a function of $\mathbf{x}_k$, where $\mathbf{w}_p\in \mathbb{C}^{M_p}$, and plug $\mathbf{w}_p$ into other implementations of neural networks with $M_p$ scalar inputs. An obvious example is the scalar spatial descriptors for predicting the interatomic potential energy \citep{BP_prl_2007} as mentioned in previous subsection, where energy has even parity symmetry.

\section{Empirical Experiments}
\label{sec:poc}

When the governing rules of a physical system are unknown, it is hard to apply any analytical method to study the evolution of a system. A viable method nowadays is to adopt an ML model supplied with a substantial amount of data. After proper training, the model will attain certain predictive power.

In atomic scale simulations, there are many developments on the interatomic potentials using different ML methods, such as Gaussian process \citep{Bartok_PRL_2010}, neural network \citep{Kondor_arxiv_2018,batzner2022, BP_prl_2007} and moment tensor \citep{Shapeev}. Atomic positions and atomic energies are used as the input and output data, respectively. The atomic energies are usually obtained from density function theory (DFT) calculations \citep{Hohenberg_Kohn,Kohn_Sham}. Atomic forces are then calculated as the derivative of the total energy (or Hamiltonian). This approach is viable only if energy can be calculated. Unfortunately, in many observations, energy is not an included quantity.   

We ask two questions here. First, can we predict forces directly from positions, without knowing the energies explicitly? This question is not limited to atomic scale modeling. We can ask similar questions in meteorology and cosmology. Second, can we predict multiple forces in a single calculation? In conventional ML interatomic potential, only one force vector is calculated from an ML model. We are going to use our ENN to demonstrate the possibility of
answering ``yes'' to both questions.

\subsection{Model and data}

We simulate a system of 4-body motion governed by a Hamiltonian model. Our aim is to predict the forces when atoms are locating at different positions, and simulate the dynamics. We generate data using a well defined physical model, which allows us to examine the errors.

\begin{figure}
    \centering
    \includegraphics[width=8cm]{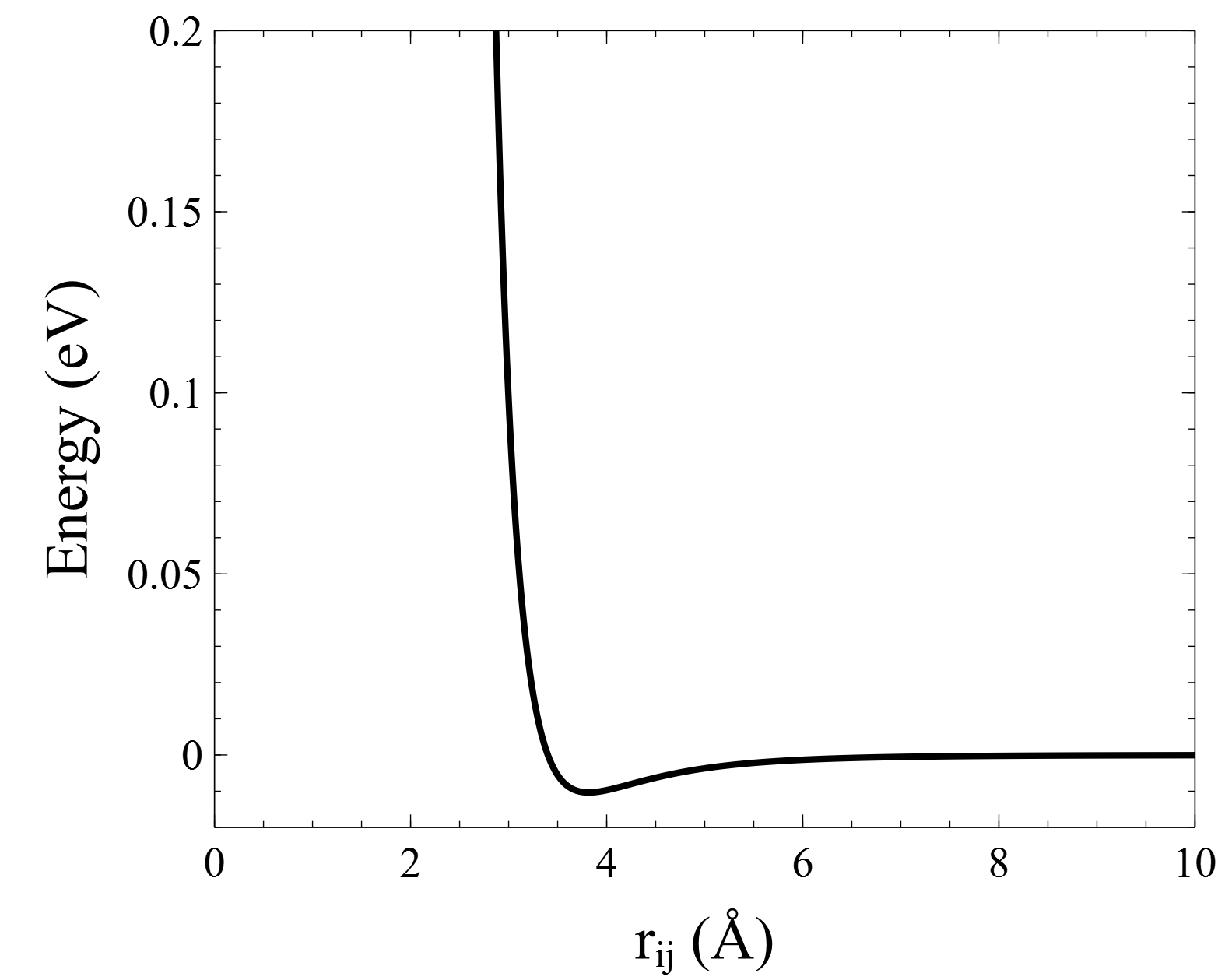}
    \caption{A plot of the Lennard-Jones potential for Argon according to equation \ref{Argon_LJ_potential}.}
    \label{fig:Argon_potential}
\end{figure}

We adopt a pair-wise Lennard-Jones potential for Argon \citep{Rahman_PhysRev_1964}:
\begin{equation}
    U_{ij}=4\epsilon\left(\left(\frac{r_0}{r_{ij}}\right)^{12} - \left(\frac{r_0}{r_{ij}}\right)^6\right),\label{Argon_LJ_potential}
\end{equation}
where $\epsilon/k_B=120$ Kelvin, $r_0=3.4${\AA}, and $k_B$ is the Boltzmann constant. A plot of the potential energy is shown in Figure~\ref{fig:Argon_potential}. 

The interatomic potential energy of the system is written as a sum of pair-wise interaction energies:
\begin{equation}
    U=\sum_{i,j,i>j}U_{ij}(r_{ij}),
\end{equation}
where $r_{ij}=|\mathbf{r}_i-\mathbf{r}_j|$. The force acting on atom $i$ is 
\begin{equation}
    \mathbf{F}_i=-\frac{\partial U}{\partial \mathbf{r}_i}.
\end{equation}

We generate 100,000 set of positions in three dimensional space. Each set contains the positions of 4 atoms. The $x$, $y$, and $z$ coordinates of each atom is generated randomly according to the Gaussian distribution with mean equals zero and standard deviation equals 3{\AA}. Then, we calculate the interatomic distance of each pair of atoms. If any of them smaller than $r_{min}=2.8${\AA}, we discard this set of positions and generate a new one. We repeat this procedure until no interatomic distance is small than $r_{min}$. This is to avoid the occurrence of very large atomic force due to small separation. We can readily understand it by inspecting Fig. \ref{fig:Argon_potential}. The energy have a drastic increase at around 3{\AA}. Atoms can hardly be in such small separation in dynamic simulations. Using these positions, we can obtain a set of four atomic forces for each set of positions using the Lennard-Jones potential. 

Instead of using the positions as inputs directly, we use the relative positions as inputs:
\begin{equation}
   \mathbf{x}_0=\{\mathbf{r}_{12}, \mathbf{r}_{13}, \mathbf{r}_{14}, \mathbf{r}_{23}, \mathbf{r}_{24}, \mathbf{r}_{34}\}. 
\end{equation}
This takes care of the translational symmetry.

The target data are simply the atomic forces:
\begin{equation}
    \mathbf{T}=\{\mathbf{F}_1,\mathbf{F}_2,\mathbf{F}_3,\mathbf{F}_4\}.
\end{equation}
We rescale both the input and output data by their standard deviations before we use them to train an ENN. Data are split, where 60\% is for training, 20\% is for validation, and 20\% for testing.

\subsection{Learning and errors}

The relationship between the input and output data is learned by an ENN, which has five hidden layers. The number of nodes in each layers counting from input to output layers are 6, 50, 90, 100, 80, 50, and 4. 

The loss function is the mean-squared error, which is defined as
\begin{equation}
    L=\frac{1}{N_{data}}\sum_{data} (\mathbf{T}-\mathbf{x}_L)^2,
\end{equation}
where $N_{data}$ is the number of used data and $\mathbf{x}_L$ is the output data. The weight parameters $\mathbf{W}$ are initialized according to normalized Xavier method \citep{Xavier}. The bias parameters $\mathbf{b}$ are initialized to zeros.

The training of ENN is performed through minimizing the loss function with respect to $\{\mathbf{W},\mathbf{b}\}$. We used the FIRE algorithm \citep{Bitzek_PRL_2006}. It is a minimization method commonly used for relaxing atomic structures. 
It is similar to the Nesterov momentum method~\citep{Nesterov1983AMF}, and has fast convergence behavior in practice.
We briefly discuss the method and our adaptation in Appendix. 

\begin{figure}
    \centering
    \includegraphics[width=8cm]{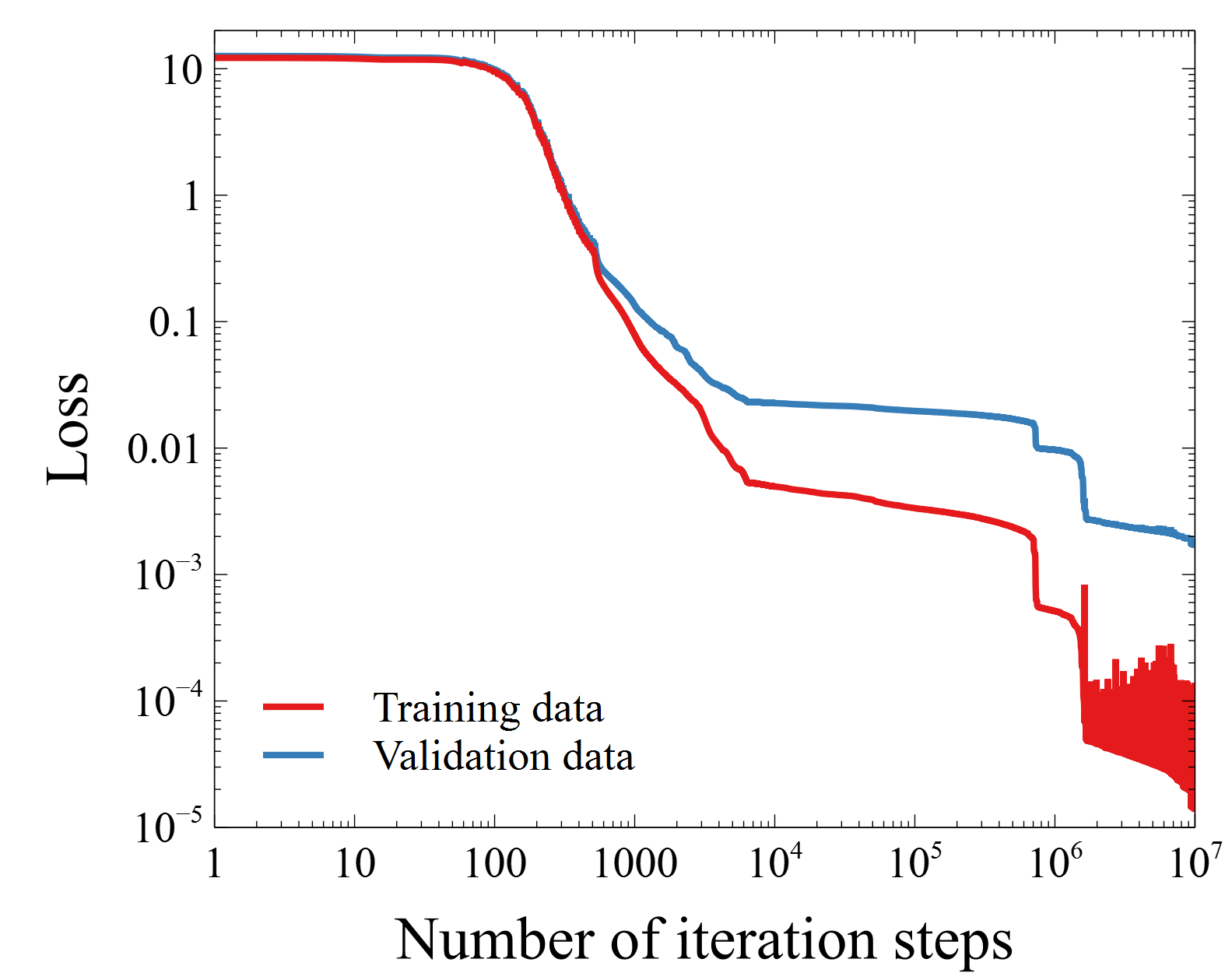}
    \caption{The value of Loss function (unitless) calculated using training data and validation data as a function of iterative steps using FIRE algorithm.} 
    \label{fig:Loss_value}
\end{figure}

Fig.~\ref{fig:Loss_value} shows the change of the value of the loss function calculated using the training data and the validation data. We performed 10 million iterations. We see that both of them drop significantly. As expected, the training loss drops faster than validation loss. However, it seems that the model does not suffer from overfitting. We stop the iteration process as soon as we observe fluctuations in the training loss, i.e., further iterations might actually increase the training loss.

\begin{figure}
    \centering
    \includegraphics[width=8cm]{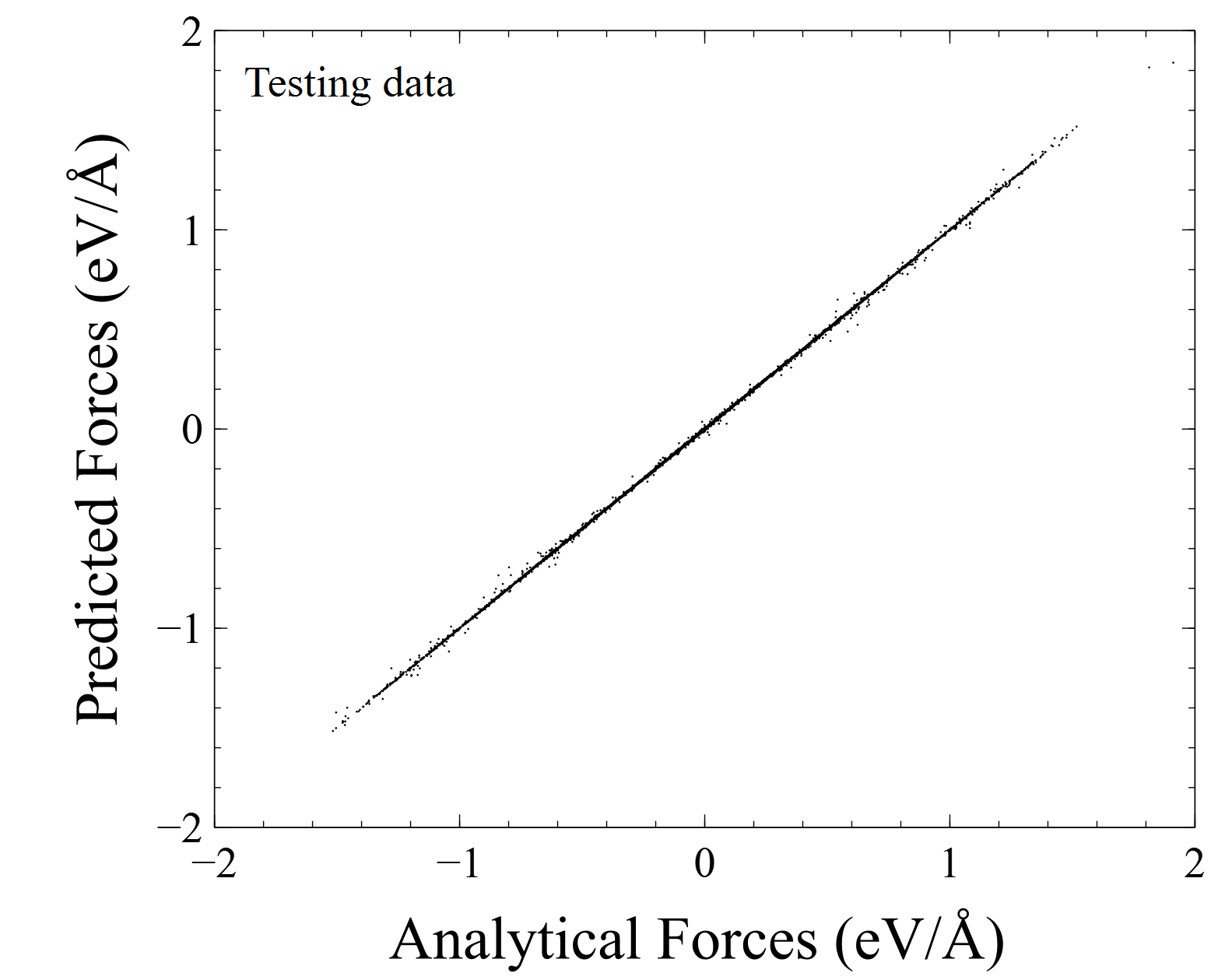}
    \caption{Each component of the atomic forces calculated analytically using the Lennard-Jones potential versus the predictions calculated using the trained ENN. They are calculated using the testing data.}
    \label{fig:Analytic_Predicted_forces}
\end{figure}

Using testing data, we can calculate the atomic forces analytically according to the Lennard-Jones potential and predict them by our trained ENN. In Fig. \ref{fig:Analytic_Predicted_forces}, the analytical and predicted values are plotted against each others. We plotted all the $x$, $y$, and $z$ components of the data. The root mean square deviation (RMSD) is 0.00118 eV/{\AA}. Observe that the training data are in the order of 0.1 to 1 eV/{\AA}, and the average error is in the order of 0.001 eV/{\AA}, which is fairly satisfactory.

\subsection{Dynamic simulations}

We use the forces predicted by our trained ENN to drive the evolution of a system of four Argon atoms using molecular dynamics (MD). We will also compare our predictions with analytical solutions. Note that our ENN was not trained to any trajectory of atomic motion. All training data are static. No history dependent information was involved in the training.

The motion of atoms are governed by the Newton's equations
\begin{eqnarray}
    \frac{d\mathbf{p}_i}{dt}&=&\mathbf{F}_i,\\
    \frac{d\mathbf{r}_i}{dt}&=&\frac{\mathbf{p}_i}{m_i},
\end{eqnarray}
where the position and momentum of atom $i$ are $\mathbf{r}_i\in\mathbb{R}^3$, $\mathbf{p}_i\in\mathbb{R}^3$ and the atomic mass is $m_i$. 

Using our trained ENN, we can predict the atomic forces $\{\mathbf{F}_i\}$. On the other hand, if the analytical form of a Hamiltonian is known, the atomic force is
\begin{equation}
    \mathbf{F}_i = -\frac{\partial \mathcal{H}}{\partial \mathbf{r}_i},
\end{equation}
where the Hamiltonian is:
\begin{equation}
    \mathcal{H}=\sum_{i} \frac{\mathbf{p_i}^2}{2m_i} + U(\{\mathbf{r}_i\}).\label{Hamiltonian}
\end{equation}
Without introducing perturbation and dissipation, this dynamic system is a closed system, and so the total energy should conserve. 

We initialized ten samples. The positions of Argon atoms are initialized at $(3,0,0.1)$, $(-3,-0.1,0)$, $(0.1,2.5,0)$, and $(0,-2.5,-0.1)$, where unit is in {\AA}. Velocities are generated randomly with kinetic energy corresponding to a temperature of 10 Kelvin. The mass of an Argon atom is 39.948$u$. We integrated Newton's equation using velocity Verlet algorithm. We used a time step of 1fs, which is a conventional value for MD simulations.

\begin{figure}
    \centering
    \includegraphics[width=8cm]{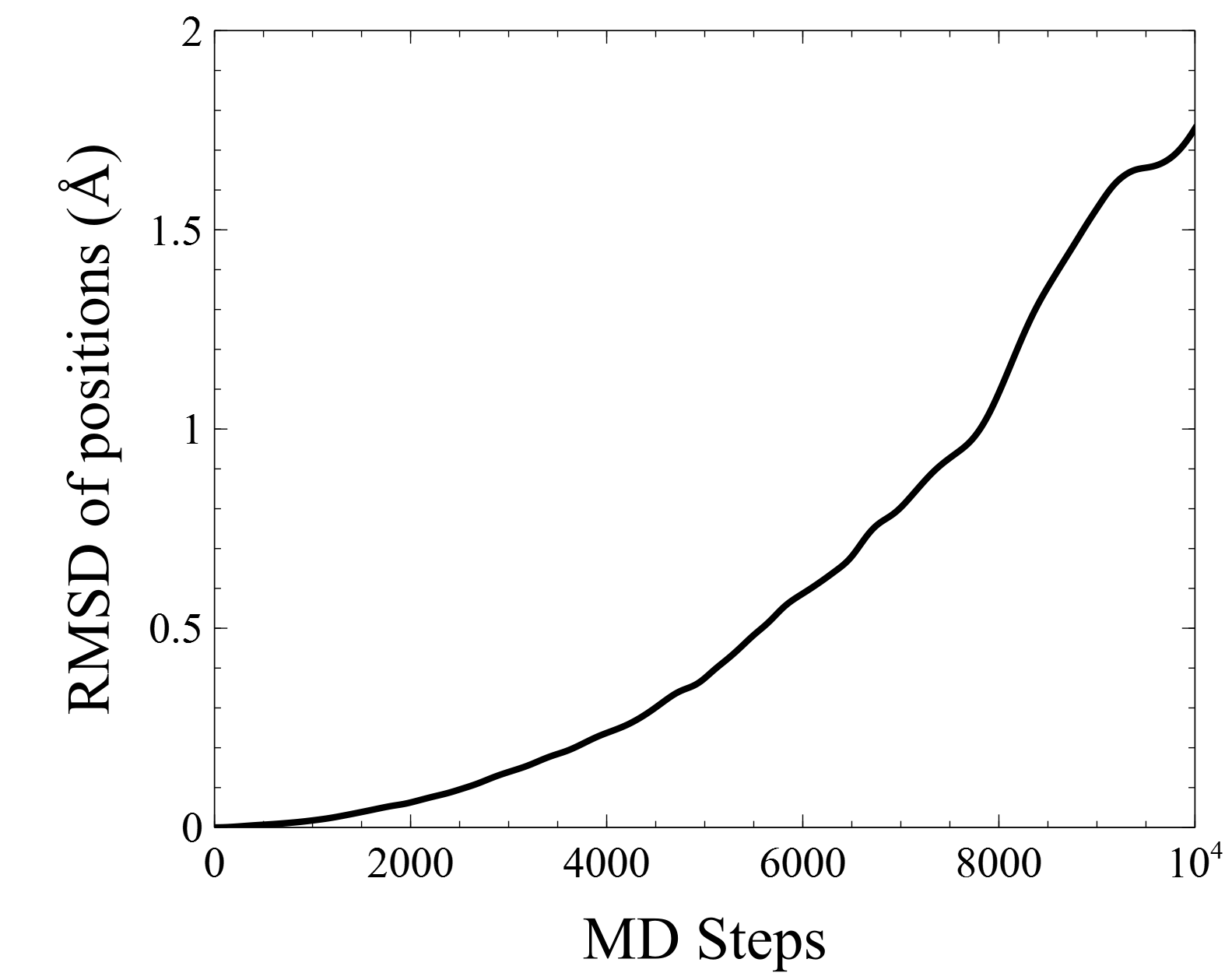}
    \caption{The root-mean-square-derivation (RMSD) of positions with respect to analytical solution and prediction by our trained ENN. The RMSD is calculated across 10 samples. Each sample contains 4 atoms.}
    \label{fig:RMSD_positions}
\end{figure}

\begin{figure}
    \centering
    \includegraphics[width=8cm]{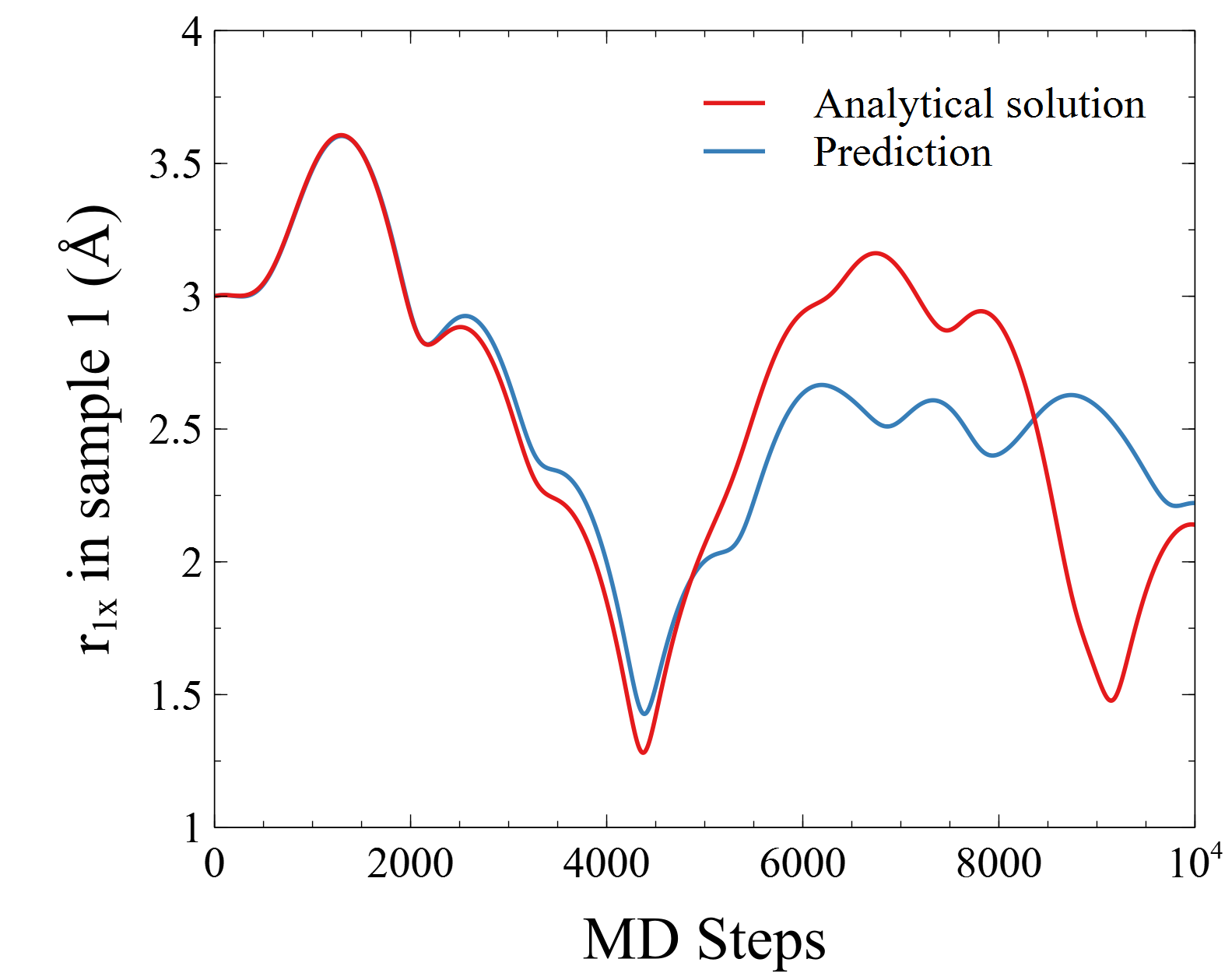}
    \caption{The $x$ component of the position of atom 1 in sample 1. Analytical solution and prediction are shown.}
    \label{fig:Position_of_r1x}
\end{figure}

We calculated the RMSD of the positions of atoms. It is defined as:
\begin{equation}
    \textrm{RMSD}(\{\mathbf{r}_i\}) = \sqrt{\frac{1}{N_{s}N_{at}}\sum_{samples,atoms} (\mathbf{r}_i^a -\mathbf{r}_i^p)^2},
\end{equation}
where $N_{s}=10$ is the number of sample, $N_{at}=4$ is the number atoms in a sample, $\mathbf{r}_i^a$ is the position of atom $i$ calculated according to analytical solution, and $\mathbf{r}_i^p$ is the position calculated using forces predicted by ENN. 

Fig. \ref{fig:RMSD_positions} shows the RMSD of positions as a function of MD steps. As expected, they deviate more and more as a function of steps, because the error is accumulating throughout the simulation. We may inspect the real trajectory of an atom in Fig. \ref{fig:Position_of_r1x}. It shows the $x$ component of atom 1 in sample 1. We see the initial 1500 MD steps predictions are fairly good, and up to 4000 MD steps are acceptable. Our ENN shows certain predictive power, and the predictions are three dimensional vectors. 

\begin{figure}
    \centering
    \includegraphics[width=8cm]{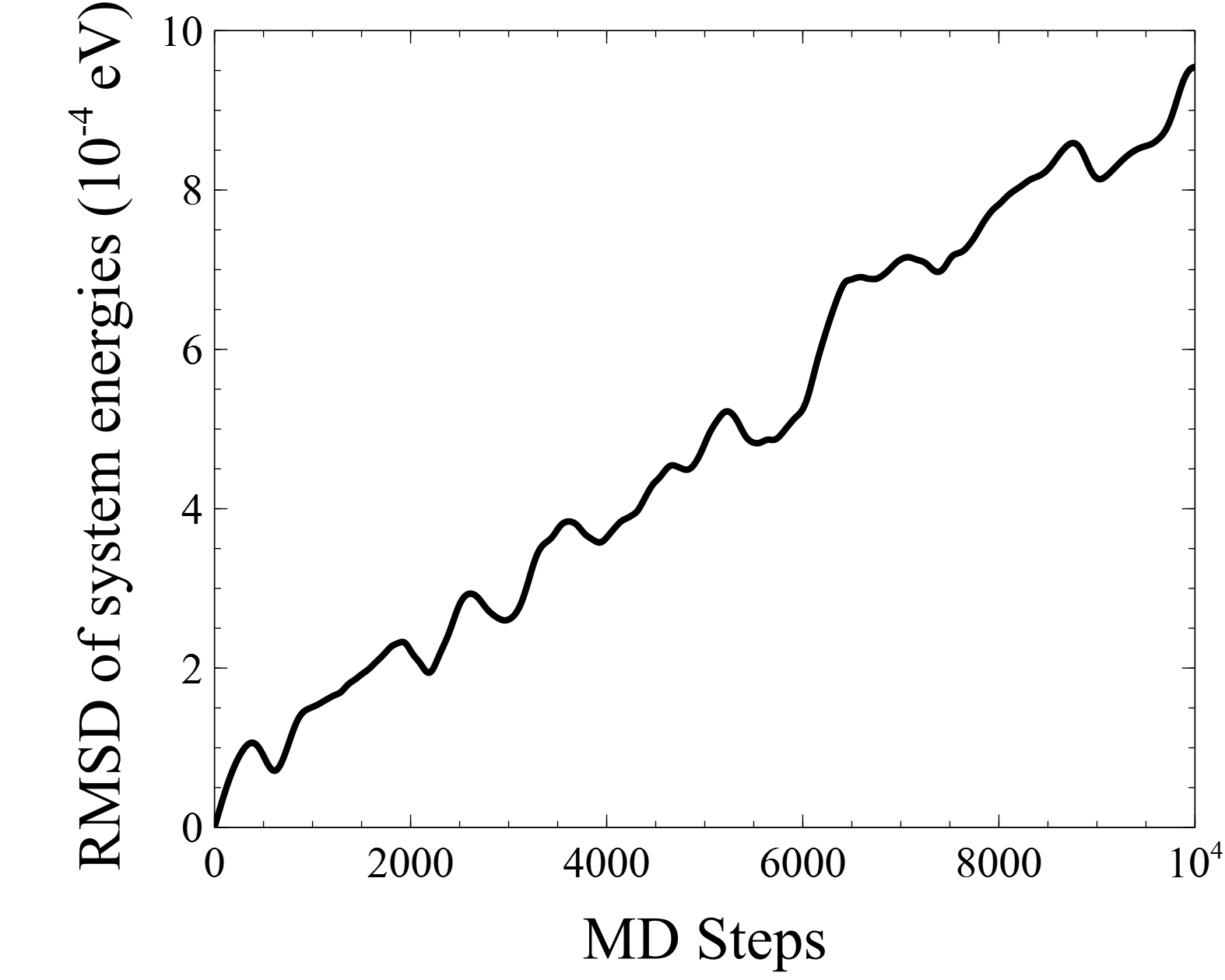}
    \caption{The RMSD of energy calculated using the positions calculated analytically or using ENN.}
    \label{fig:RMSD_energy}
\end{figure}

\begin{figure}
    \centering
    \includegraphics[width=8cm]{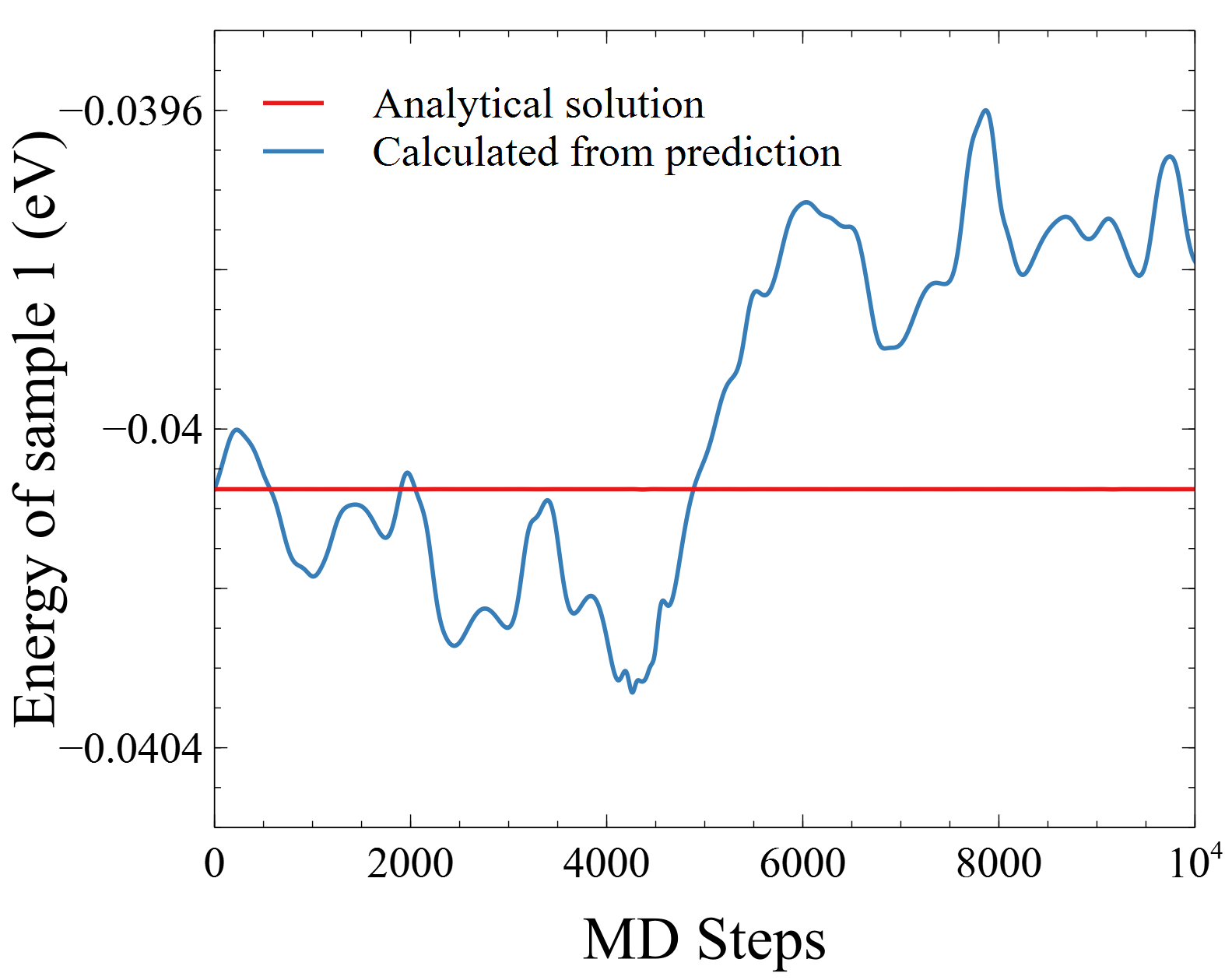}
    \caption{The system energy of sample 1. analytical solution and prediction are shown.}
    \label{fig:energy_sample1}
\end{figure}

We calculated the RMSD of the system energies. It is defined as:
\begin{equation}
    \textrm{RMSD}(E) = \sqrt{\frac{1}{N_{s}}\sum_{samples} (E^a -E^p)^2},
\end{equation}
where $E^a$ and $E^p$ are the total energy calculated using equation \ref{Hamiltonian}. The potential energy are calculated using the positions of atoms that evolve according to forces calculated analytically or by prediction using ENN.

Fig. \ref{fig:RMSD_energy} shows the RMSD of energy across ten samples. Again, we can see the deviation accumulates. However, we should remember that in the training of ENN, we did not provide any information about energy to the training. If we look at around 4000 steps, the RMSD is less than about $4\times 10^{-4}$ eV, which is about 2 order of magnitude smaller than the system energy. Our results are encouraging. It shows that even if we don't know the system total energy (or Hamiltonian), we can still predict forces, which are vectors, using our ENN. We can also predict multiple forces at the same time. 

\section{Conclusion}
We have designed a new feedfoward ENN for unitary transformations. It does not involve convolution with higher order representation, such as spherical harmonics and Wigner matrices. Moreover, our model works for vectors in arbitrary dimensions. Our ENNs can be trained by efficient backpropagation and an extra layer of GNN can be added to achieve permutation symmetry. An example on the dynamics of Argon atoms is given showing the practicality of our architecture via empirical simulations.

\section*{Declaration of competing interest}
The authors declare that they have no known competing financial interests or personal relationships that could have appeared to influence the work reported in this paper.

\section*{Acknowledgments}
This work has been carried out within the framework of the EUROfusion Consortium, funded by the European Union via the Euratom Research and Training Programme (Grant Agreement No 101052200 — EUROfusion) and from the EPSRC [grant number EP/W006839/1].  To obtain further information on the data and models underlying this paper please contact PublicationsManager@ukaea.uk. Views and opinions expressed are however those of the author(s) only and do not necessarily reflect those of the European Union or the European Commission. Neither the European Union nor the European Commission can be held responsible for them.
T.-H. Hubert Chan was partially supported by the Hong Kong RGC under the grants 17201220 and 17202121.

\appendix
\section{FIRE minimization algorithm}
FIRE (fast inertial relaxation engine) \citep{Bitzek_PRL_2006} is a minimization algorithm commonly used in atomic scale simulations for structural relaxation. We are going to briefly mention the algorithm below and discuss our adaptation.

Assuming we have a system of atoms governed by the Hamiltonian $\mathcal{H}$, we may find a configuration with potential energy at local minimum through following steps.

Step 1: Define parameters $N_{min}$, $f_{inc}$, $f_{dec}$, $\alpha_{start}$, $f_\alpha$, $\Delta t$, $\Delta t_{max}$, and $i_{max}$. Set $\alpha = \alpha_{start}$, $N=0$, and $i=0$.

Step 2: Set the initial positions $\mathbf{x}$ and atomic mass $m$. Initialize velocities $\mathbf{v}=0$.

Step 3: Calculate the atomic forces $\mathbf{F}=-\nabla \mathcal{H}(\mathbf{x})$.

Step 4: Put 
\begin{eqnarray}
\mathbf{x}(t + \Delta t) &=& \mathbf{x}(t) + \mathbf{v}\Delta t, \nonumber\\
\mathbf{v}(t + \Delta t) &=& \mathbf{v}(t) + \frac{\mathbf{F}}{m}\Delta t. \nonumber
\end{eqnarray}

Step 5: Calculate $P=\mathbf{F}\cdot \mathbf{v}$.

Step 6: Put $N\rightarrow N+1$ and set
\begin{equation}
\mathbf{v} \rightarrow (1-\alpha)\mathbf{v}  + \alpha |\mathbf{v}|\frac{\mathbf{F}}{|\mathbf{F}|}. \nonumber
\end{equation}

Step 7: if $P > 0$ and $N > N_{min}$, set 
\begin{eqnarray}
    \Delta t &\rightarrow& \min (\Delta t f_{inc}, \Delta t_{max})\nonumber\\
    \alpha &\rightarrow& \alpha f_{\alpha}\nonumber
\end{eqnarray}

Step 8: if $P \leq 0$, set 
\begin{eqnarray}
    \Delta t&\rightarrow& \Delta t f_{dec}\nonumber\\
    \mathbf{v}&\rightarrow&\mathbf{0}\nonumber\\
    \alpha &\rightarrow& \alpha_{start}\nonumber\\
    N&\rightarrow& 0\nonumber
\end{eqnarray}

Step 9: Set $i\rightarrow i+1$. Go to Step 3, or end if $i > i_{max}$.

In our case, we are minimising the Loss function with respect to the weight and bias parameters $\{\mathbf{W}, \mathbf{b}\}$. We flattened $\{\mathbf{W}, \mathbf{b}\}$ to a column vector and treated it as $\mathbf{x}$. We also flattened the gradient of the Loss function and treated it as the negative of $\mathbf{F}$. After some trials and errors, we used a pseudo mass $m=0.1$, $\Delta t= 0.001$, and $\Delta t_{max} = 0.01$. For other parameters, we follow the original suggestions \citep{Bitzek_PRL_2006}, $N_{min}=5$, $f_{inc}=1.1$, $f_{dec}=0.5$, $\alpha_{start}=0.1$, and $f_\alpha =0.99$.

\bibliographystyle{elsarticle-harv} 
\bibliography{reference}





\end{document}